\documentclass{article}

\usepackage[final]{arxiv}

\usepackage[utf8]{inputenc} 
\usepackage[T1]{fontenc}    
\usepackage{hyperref}       
\hypersetup{
    colorlinks=true,
    linkcolor=blue,    
    citecolor=blue,     
    urlcolor=blue      
}
\usepackage{url}            
\usepackage{booktabs}       
\usepackage{amsfonts}       
\usepackage{nicefrac}       
\usepackage{microtype}      
\usepackage{xcolor}         
\usepackage{graphicx}
\usepackage{multirow}
\usepackage{pifont}
\usepackage{colortbl}
\usepackage{subcaption}
\usepackage{lipsum} 
\usepackage{natbib}
\usepackage{wrapfig}
\usepackage{amsmath}
\usepackage{amssymb}
\bibpunct{[}{]}{,}{n}{,}{,}

\usepackage {ulem}

\newcommand{\etal}{\textit{et al.}}
\title{Diff-Privacy: Diffusion-based Face Privacy Protection}

\author{
\qquad Xiao He\textsuperscript{1}
\qquad Mingrui Zhu\textsuperscript{1}
\qquad Dongxin Chen\textsuperscript{1}
\qquad {Nannan Wang\textsuperscript{1, $\dagger$}}
\qquad \textbf{Xinbo Gao\textsuperscript{2}}
\\
\textsuperscript{1}Xidian University\qquad \\ \textsuperscript{2}Chongqing University of Posts and Telecommunications\\ 
{\tt\small {nnwang}@xidian.edu.cn}\\
}

\begin{document}

\maketitle

\begin{abstract}
Privacy protection has become a top priority as the proliferation of AI techniques has led to widespread collection and misuse of personal data. Anonymization and visual identity information hiding are two important facial privacy protection tasks that aim to remove identification characteristics from facial images at the human perception level. However, they have a significant difference in that the former aims to prevent the machine from recognizing correctly, while the latter needs to ensure the accuracy of machine recognition. Therefore, it is difficult to train a model to complete these two tasks simultaneously. In this paper, we unify the task of anonymization and visual identity information hiding and propose a novel face privacy protection method based on diffusion models, dubbed Diff-Privacy. Specifically, we train our proposed multi-scale image inversion module (MSI) to obtain a set of SDM format conditional embeddings of the original image. Based on the conditional embeddings, we design corresponding embedding scheduling strategies and construct different energy functions during the denoising process to achieve anonymization and visual identity information hiding. Extensive experiments have been conducted to validate the effectiveness of our proposed framework in protecting facial privacy.
\end{abstract}


\section{Introduction}
\label{sec:introduction}
The widespread application of intelligent algorithms and devices brings convenience together with security concerns. Personal images uploaded on social media platforms or captured through intelligent surveillance systems can be vulnerable to unauthorized access and misuse by malicious actors, posing a significant threat to personal privacy. On the one hand, we are eager to use technology to improve our quality of life (such as video conferencing), but on the other hand, we are unwilling to give up our personal privacy. Consequently, research in the field of privacy protection has garnered significant importance, with a particular emphasis on safeguarding facial images that contain substantial amounts of sensitive information. 

Recent face privacy protection methods can be broadly divided into two main categories, $i.e.,$ anonymization \cite{neustaedter2006blur,newton2005preserving,gross2009face,hukkelaas2019deepprivacy,maximov2020ciagan,cao2021personalized,gu2020password,li2023riddle} and visual identity information hiding \cite{ito2021image,Su2022VisualIH}. Anonymization methods aim to remove identification characteristics from images while retaining essential facial structures to ensure the functionality of face detection algorithms. Crucially, anonymized faces should maintain a realistic appearance while preventing human observers and facial recognition models from recognizing their identities correctly. Unlike anonymization, face images processed by visual identity information hiding methods are unrecognizable to human observers but can be recognized by machines. Regarding application scenarios, the former (anonymization) allows people to share photos with anonymized faces on public social media. The latter (visual identity information hiding) encrypts private images stored in cyberspace, ensuring the accuracy of facial recognition functions while improving security.

\begin{figure}
\centering
\includegraphics[width=1.0\linewidth]{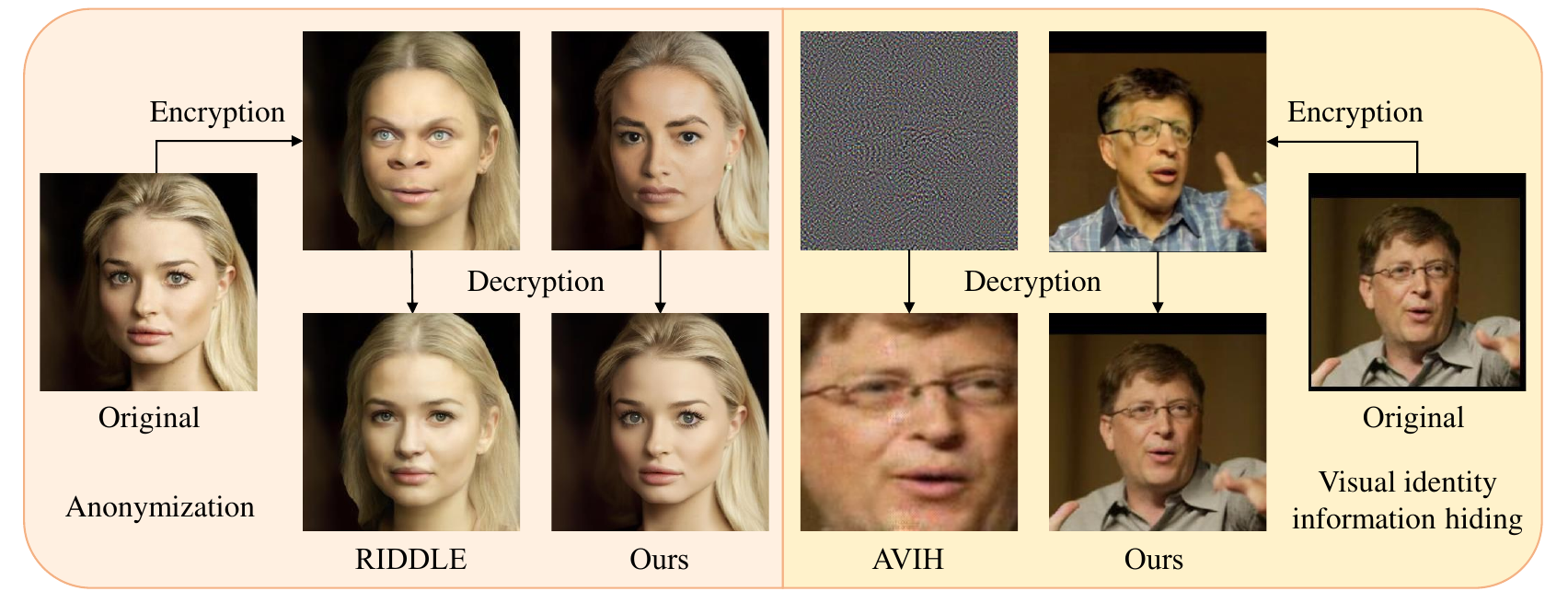}
\caption{We show the results of our method in anonymization and visual identity information hiding tasks. Compared with existing methods \cite{li2023riddle,Su2022VisualIH}, our method can better generate photo-realism encrypted images and recover the original image.}
\label{fig:intro}
\vspace{-2mm}
\end{figure}

However, the above technologies for facial privacy protection often specialize in specific types of protection and rely on high-quality facial datasets for training or continuous online iteration of images. Moreover, these technologies often leave noticeable editing traces and possess limited recovery capabilities. Consequently, there is an urgent need to develop a method that can effectively and flexibly achieve facial privacy protection for various requirements, mitigating the shortcomings of existing approaches. However, due to the importance placed on data privacy by relevant institutions, collecting high-quality facial datasets is extremely challenging. For example, the European Union has enacted legislation such as the General Data Protection Regulation (GDPR) to safeguard personal data, and a popular personal re-identification dataset, Duke MTMC, has been taken offline due to privacy reasons. Therefore, it is necessary to develop methods to protect facial privacy with only a few images effectively. Encouraged by the powerful generation ability of diffusion models, we utilize pre-trained diffusion models to promote facial privacy protection.

In this paper, we unify the task of anonymization and visual identity information hiding and propose a novel face privacy protection method, dubbed Diff-Privacy. Diff-Privacy enables flexible facial privacy protection and ensures identity recovery when needed. As shown in Figure.~\ref{fig:intro}, Diff-Privacy generates highly realistic faces whose identities differ from the original faces during the task of anonymization and visual identity information hiding. Furthermore, Diff-Privacy also exhibits exceptional recovery quality and some other advantages. These include: \textbf{Security.} Given an encrypted image, the original image is recovered only if the correct password is provided. \textbf{Utility.} The encrypted images generated by our method can still be used for downstream computer vision tasks, such as face detection. \textbf{Diversity.} Given an original image, our model can generate a series of encrypted images that are different from each other. \textbf{Controllability.} Our method can achieve face privacy protection while identity-independent attributes such as background and posture remain unchanged. We have conducted extensive experiments on publicly available facial datasets to verify the effectiveness of our method.

In summary, we make the following contributions:
\begin{itemize}
    \item We unify the task of anonymization and visual identity information hiding and propose a novel diffusion-based face privacy protection method, which can effectively achieve facial privacy protection and identity recovery.
    \item We develop an energy function-based identity guidance module to perform gradient correction on the denoising process to ensure the correctly or incorrectly recognized by machines of different privacy protection tasks. We enhance the diversity of anonymized images by maximizing the difference of face identities generated under different random noises.
    \item According to the characteristics of the Diffusion model that different time steps pay attention to different-level information, we design a multi-scale image inversion module to learn conditional embedding and propose corresponding embedding scheduling strategies to meet different privacy protection requirements.
    \item Experimental results show that compared with existing methods, our method can significantly change facial identity visually while maintaining its photo-realism, in addition to high-quality recovery results.
\end{itemize}

\section{Related work}
\label{sec:related_work}

\subsection{Anonymization}
Anonymization aims to remove human recognition features from an image, rendering it unrecognizable to both human observers and computer vision systems. Existing anonymization methods can be categorized into two main types based on the underlying technology: low-level image processing methods \cite{boyle2000effects,chen2007tools,neustaedter2006blur,tansuriyavong2001privacy,newton2005preserving,gross2009face} and face replacement-based methods \cite{hukkelaas2019deepprivacy,gafni2019live,sun2018hybrid,sun2018natural,maximov2020ciagan,cao2021personalized,gu2020password,li2023riddle}. The first category employs techniques such as blurring, mosaicing, masking and pixelization to achieve face anonymization. Although these techniques can completely eliminate identity, they can also seriously damage the utility of the original image. To address this limitation, researchers have leveraged the capabilities of Generative Adversarial Networks (GANs) to tackle the challenge of face anonymization. These methods focus on generating a virtual face to replace the original identity, known as face replacement-based methods. Sun~\etal~\cite{sun2018natural} tries to mask the face region in the image and then generates a new face by inpainting. However, the anonymized faces generated by the above methods often have similar and unnatural appearances. In addition, Maximov~\etal~\cite{maximov2020ciagan} forces the generated image to display different identities with similar features to the source image based on landmark information and occluded images. It can generate a variety of encrypted images for the original image, but for high-resolution images, it still lacks the ability to generate natural faces. 


Recently, researchers have focused on the development of recoverable anonymization methods \cite{cao2021personalized,gu2020password,li2023riddle}. Gu~\etal~\cite{gu2020password} trains a conditional GAN with multi-task learning objectives, which takes the input image and password as conditions and outputs the corresponding anonymization image.   Cao~\etal~\cite{cao2021personalized} decouples a face image into an attribute vector and identity vector and rotates the identity vector to change identity. Li~\etal~\cite{li2023riddle} projects the original image into the latent space of the pre-trained StyleGAN2 and processes the latent code and password through a lightweight transformer to generate encrypted code. However, these methods often rely on high-quality facial datasets for training and can not achieve satisfactory results in terms of the quality of anonymized and recovered images.

\subsection{Visual information hiding}
Visual information hiding focuses on the human visual perspective, aiming to encrypt the source image so that human observers cannot recognize it. It mainly includes Homomorphic encryption (HE)-based methods \cite{aono2017privacy,liu2015deep,wang2018efficient} and perceptual encryption (PE)-based methods \cite{ding2020deepedn,ito2021image,sirichotedumrong2019privacy,sirichotedumrong2021gan}. HE-based methods mainly come from Cryptography and are usually unsuitable for deep neural networks (DNNs) containing many nonlinear operations. For PE-based methods, some work \cite{ding2020deepedn,sirichotedumrong2019privacy,sirichotedumrong2021gan} focuses on designing or finding the encrypted domain and directly using the encrypted images to train the model. However, this training strategy has a significant impact on the accuracy of the model. To address this issue, Ito~\etal~\cite{ito2021image} trains a transformation network that aims to preserve the correct classification results of the classifier while hiding visual information. However, a limitation of this method is that the protected image cannot be recovered to its original form. Inspired by adversarial attack methods, Su~\etal~\cite{Su2022VisualIH} proposes a visual identity information hiding method to protect facial privacy protection. It utilizes recovery and recognition models to continuously iterate images through Type-I attacks, which achieves image encryption and recovery. However, the images generated by this method are similar to random noise, making it easy for hackers to realize that these images are encrypted. In addition, this online optimization-based method leads to slower generation speed.

\subsection{Diffusion models}
Diffusion-based Generative models (DMs) \cite{song2019generative,song2020score,ho2020denoising,sohl2015deep,yang2022diffusion} is a powerful tool for complex Data modeling and generation, which has achieved the first results in density estimation and sample quality. Early work \cite{ho2020denoising} relied on Markov chains and required many iterations to generate high-quality samples. DDIM \cite{song2020denoising} proposes a deterministic sampling process that greatly reduces the time required to generate samples. Dhariwal~\etal~\cite{dhariwal2021diffusion} proposes introducing category information into the Diffusion model, which can better generate realistic images. However, this method requires to train an additional classifier, resulting in a high training cost. Classifier-free diffusion method \cite{ho2022classifier} jointly trains the conditional Diffusion model and the unconditional Diffusion model, and combines the noise estimation of the two models to achieve the balance between sample quality and diversity. In addition, considering the drawbacks of slow training and inference in pixel space of the above methods, Rombach~\etal~\cite{rombach2022high} proposes to denoise in the latent space of the pre-trained autoencoder, significantly reducing the diffusion model's computational requirements. So far, the diffusion model has been applied to various computer vision tasks.

\begin{figure}[t]
\centering
\includegraphics[width=1.0\linewidth]{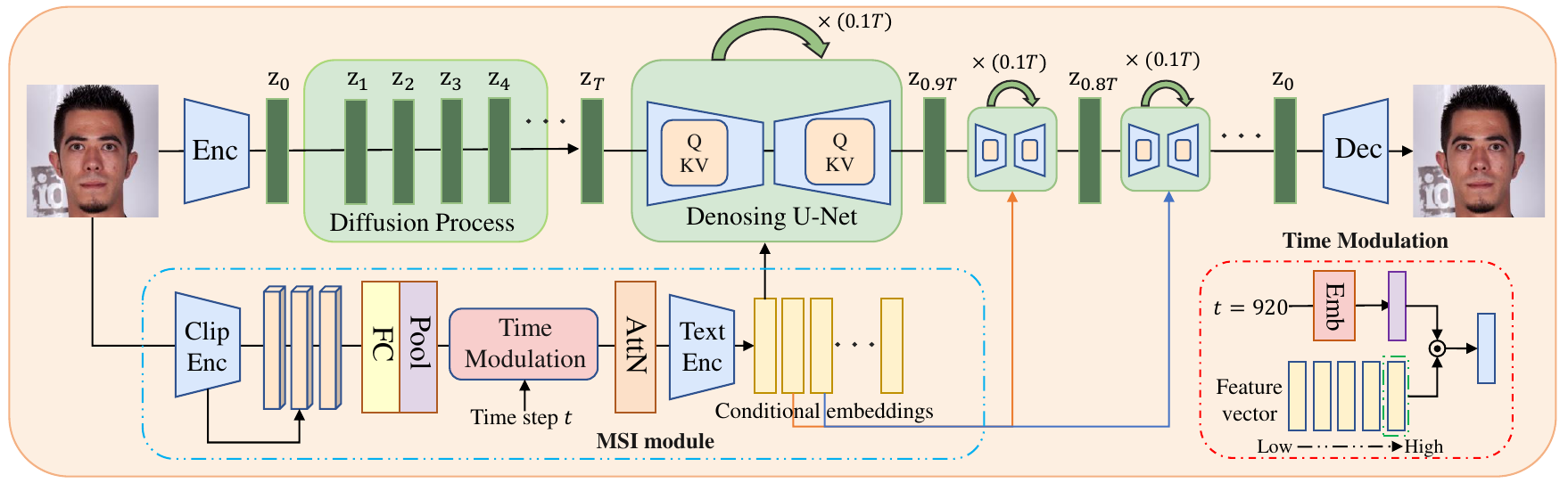}
\caption{\textbf{Training process of Diff-Privacy.} We apply the stable diffusion model (SDM) as the generative backbone and propose a multi-scale image inversion module. During the training process, the parameters of SDM are fixed. We only optimize our MSI module to extract a set of conditional embeddings.}
\label{fig:method_train}
\end{figure}

\section{Diff-Privacy}
\label{sec:method}
To fulfill the goals we mentioned in Section \ref{sec:introduction}, we introduce Diff-Privacy, a diffusion-based face privacy protection method. One notable advantage of Diff-Privacy is its inherent flexibility in achieving facial privacy protection to meet diverse requirements. Specifically, the main framework of the method can be summarized into the following three stages. \textbf{Stage I.} Learning the corresponding conditional embedding of the image in the pre-trained SDM as the key-E. \textbf{Stage II.} Accomplishing privacy protection through our energy function-based identity guidance module and embedding scheduling strategy during the denoising process and then getting a noised map as the key-I according to DDIM inversion. \textbf{Stage III.} Performing identity recovery using DDIM sampling based on the acquired key. We will first briefly review some preliminaries in Section~\ref{sec:Preliminaries}. In Section~\ref{subsec:Conditional embedding learning}, We provide the training process for learning the conditional embedding of the original image. Based on the learned conditional embedding, we develop an energy function-based identity guidance module and an embedding scheduling strategy to achieve face privacy protection in Section~\ref{sec:privacy protection}. Last, we describe the details of how to recover the original face identity in Section~\ref{sec:recover}.

\subsection{Preliminaries}
\label{sec:Preliminaries}
Diffusion Models \cite{ho2020denoising} are probabilistic models designed to learn a data distribution $p(x)$ by gradually denoising a normally distributed variable, which corresponds to learning the reverse process of a fixed markov Chain of length $T$. In this paper, we focus on the pre-trained SDM, which is widely used in recent research. SDM first encodes the image $x_0$ into latent space with a pre-trained encoder $E_{nc}$, $i.e., z_0 = E_{nc}(x_0)$. It then performs the noising and denoising processes in the latent space. Finally, the pre-trained decoder $D_{ec}$ takes latent code as input and outputs the desired image. Specifically, the noising process refers to the process of gradually adding gaussian noise to the original data $z_0$ until the data becomes random noise $z_T$, which is known as a fixed-length markov chain. An important property of this process is that we can directly sample the latent code $z_t$ at any step $t\in\left\{0,..., T\right\}$ based on the original data $z_0$:
\begin{equation}
    z_t = \sqrt{\alpha_t} z_0 + \sqrt{1-\alpha_t} \epsilon,
    \label{eq:one_step nosing}
\end{equation}
where $\alpha_t = \prod_{i=1}^{t} (1-\beta_i)$. $\beta_i \in (0,1), \beta_1 < \beta_2<...<\beta_T$. Regarding the reverse process, since we aim to recover the original image accurately, we employ the deterministic DDIM sampling \cite{song2020denoising}:
\begin{equation}
    z_{t-1}= \sqrt{\frac{\alpha_{t-1}}{\alpha_t}}z_t + \left(\sqrt{\frac{1}{\alpha_{t-1}}-1}-\sqrt{\frac{1}{\alpha_t}-1}\right)\cdot \epsilon_\theta\left(z_t,t,C\right),
    \label{eq:denosing}
\end{equation}
where $C$ is the conditional embedding and $\epsilon_ \theta$ is a time-conditioned UNet equipped with attention mechanism and trained to achieve the objective. The training objectives are as follows:
\begin{equation}
    min_\theta E_{z_0,\epsilon~N(0,I),t} {\left ||\epsilon-\epsilon_\theta\left(z_t,t,C\right)\right ||}_2^2.
    \label{eq:sdm loss}
\end{equation}
\textbf{DDIM inversion.} A simple inversion technique was suggested for the DDIM sampling \cite{song2020denoising}, based on the assumption that the ODE process can be reversed in the limit of small steps:
\begin{equation}
    z_{t+1}= \sqrt{\frac{\alpha_{t+1}}{\alpha_t}}z_t + \left(\sqrt{\frac{1}{\alpha_{t+1}}-1}-\sqrt{\frac{1}{\alpha_t}-1}\right)\cdot \epsilon_\theta\left(z_t,t,C\right).
    \label{eq:nosing}
\end{equation}
Through this diffusion process, we acquire a noise map that is amenable to restoration back to the source image.

\subsection{Conditional embedding learning}
\label{subsec:Conditional embedding learning}
Inspired by \cite{daras2022multiresolution,zhang2023prospect,balaji2022ediffi}, we note that the generation process of the diffusion model is related to the frequency of the corresponding attribute's signal. Generally, the model tends to generate the overall layout at the initial stage of the denoising process (corresponding to a large time step), the structure and content at the intermediate stage, and the detailed texture at the final stage. Based on this observation, our key insight is whether we can learn a set of conditional embeddings of a specific image and design scheduling strategies for embeddings to help achieve different privacy protection tasks.

\begin{figure}[t]
\centering
\includegraphics[width=1.0\linewidth]{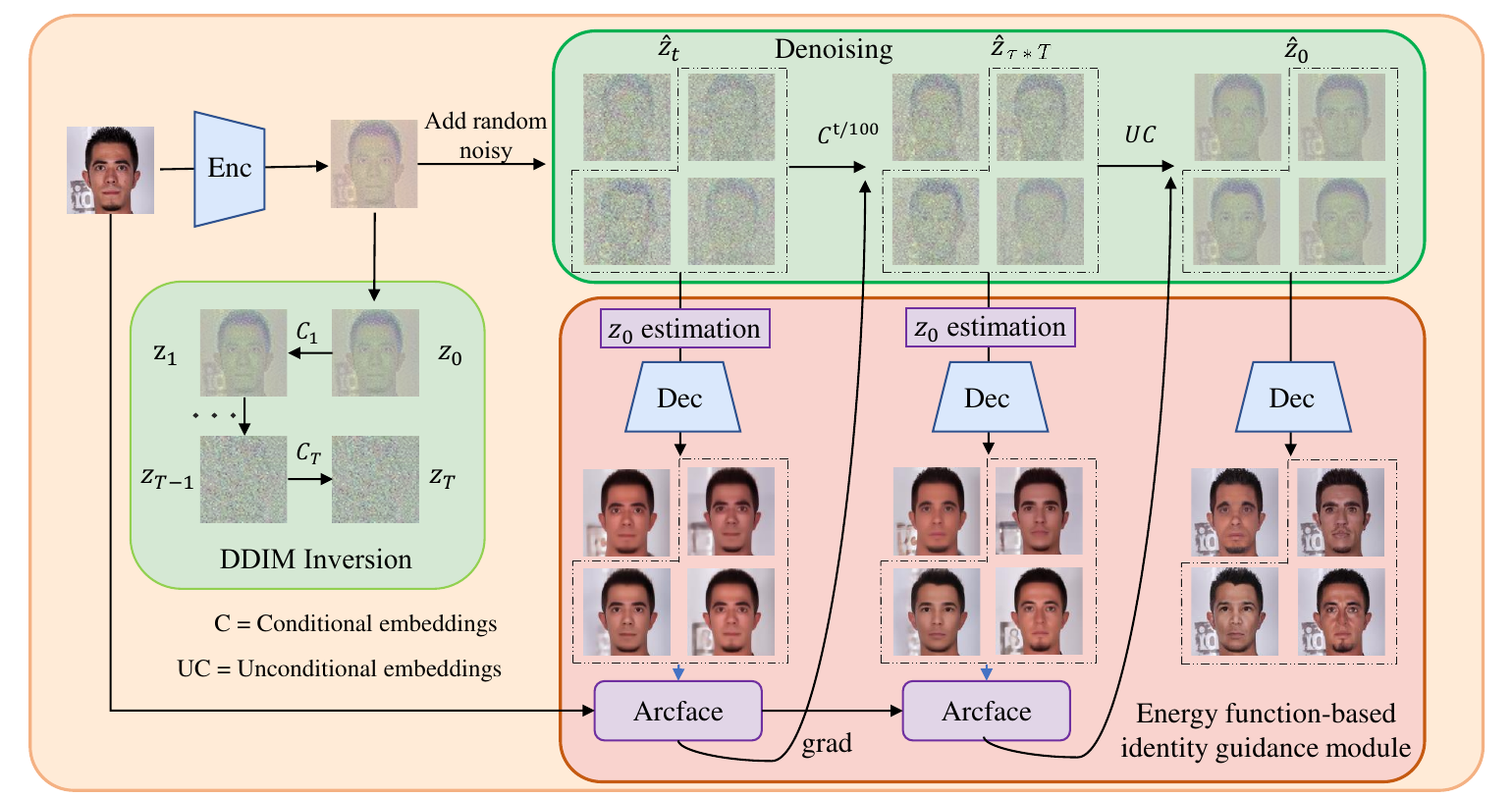}
\caption{\textbf{Inference pipeline of Diff-Privacy.} We take the anonymization task as an example to provide the inference process. The content enclosed within the black dashed box in the figure signifies its exclusive relevance to the anonymization task. It introduces four different groups of noise to each clean latent code $z_0$ and constrains the noised results $\hat{z}_k$ to enhance the diversity of anonymized images.
}
\label{fig:method_inference}
\end{figure}

We define the conditional embedding space as $C^*$. $C^*$ is an extension of the previous conditional embedding space, which provides new insight into the generation process of the diffusion model from the perspective of steps. It does not view the diffusion process as a whole but rather as different stages of generation, each stage corresponding to a unique condition embedding, generating corresponding attributes. Specifically, we divide the 1000 steps of conditioning in SDM into 10 stages on average. 
\begin{equation}
    C^* = \left\{C^0,C^1...,C^9 \right\},
    \label{eq:embedding space}
\end{equation}
where $C^{i}$ represents the token embedding corresponding to the conditional prompt in the $i-th$ stage of the diffusion model. This set of conditional embeddings is located in the CLIP text image space, and its size is set to $N$ $\times$ 768 ($N$=10 represents the number of stages). This partitioning strategy aims to maintain a balance between speed and controllability. Next, we will introduce the method of learning embedding $C^*$ from a single or a few images based on this embedding space.

An instinctive way to obtain this set of embeddings $C^i$ is to directly optimize it by minimizing the SDM loss of a specific image. However, this optimization-based approach is inefficient, and it is difficult to obtain accurate embeddings without overfitting with a single image as training data \cite{zhang2023inversion}. Thus, we design a multi-scale image inversion (MSI) module to learn conditional embeddings.

The aligned latent space of image and text embedding in CLIP provides guidance for our optimization process. However, our diffusion model uses different conditional embeddings to generate corresponding attributes at different stages. Only using the last layer features encoded by CLIP to obtain embedding will result in entangled word embedding and poor editing. In order to obtain more information and high-editable embeddings, the MSI module utilizes five layers of features from the CLIP image encoder $\tau_\theta$ and maps each layer to a 768-dimensional vector. After obtaining five 768-dimensional vectors, we modulated the vectors through our designed time embedding layer. The time embedding layer maps time steps into time embedding and performs point multiplication with corresponding feature vectors. (Note: Deep feature vectors correspond to time embeddings mapped by large time step; for example, the fifth layer corresponds to two sets of time embeddings: time steps 800-900 and 900-1000.) Subsequently, the MSI module executes attention on these image embeddings, extracting pivotal information and transmitting it to the text encoder \cite{radford2021learning} for obtaining the final embeddings.

\begin{equation}
   {C}^* = MSI(\tau_\theta(x)).
    \label{eq:MSI}
\end{equation}

We optimize MSI module by minimizing the SDM loss function. The overall training process is illustrated in Figure \ref{fig:method_train}. To avoid overfitting, we apply a dropout strategy in each cross-attention layer, which is set to 0.05. Our optimization goal can finally be defined as:

\begin{equation}
   L_{SDM} = E_{z,x,t}\left[\left ||\epsilon-\epsilon_\theta\left(z_t,t,C_t\right)\right||_2^2\right], C_t=C^{t/100}
    \label{eq:loss}
\end{equation}
where $x$ denotes the image and $C_t$ denotes the embedding used in time step $t$. $z\sim E(x),\epsilon \sim N(0,1)$, $\epsilon_\theta$ and $\tau_\theta$ are fixed during training.

\subsection{Face privacy protection}
\label{sec:privacy protection}
To achieve privacy protection, we construct an energy function and design embedding scheduling strategies to guide the denoising direction. Figure \ref{fig:method_inference} shows an overview of the proposed framework, which can achieve face anonymization and visual identity information hiding. In the subsequent sections, we will present the embedding scheduling strategy in Section \ref{subsec:Embedding scheduling strategy} and the energy function-based identity guidance module in Section \ref{subsec:Energy function-based identity guidance module}. Based on these two main components, we will introduce the implementation of face anonymization in Section \ref{subsec:anonymization} and the implementation of visual identity information hiding in Section \ref{subsec:Visual information hiding}.

\subsubsection{Embedding scheduling strategy}
\label{subsec:Embedding scheduling strategy}
As mentioned in Section~\ref{subsec:Conditional embedding learning}, we learn a set of conditional embeddings of a specific image and apply them to different stages of the denoising process. Generally, diffusion models generate images in the order of ``layout $\rightarrow$ content/structure $\rightarrow$ texture/style''. Based on this characteristic of the diffusion model, we can design an embedding scheduling strategy that facilitates privacy protection. For example, we can use the learned condition embedding in the initial stage of denoising to ensure the manifold structure of human faces and use unconditional embedding in the middle and later stages to generate different appearances of faces. We will introduce our embedding scheduling strategies designed for different privacy protection tasks later.

\subsubsection{Energy function-based identity guidance module}
\label{subsec:Energy function-based identity guidance module}
 One pivotal concern in privacy protection is the recognition rate of faces in images by machines. In order to generate images that can be correctly or incorrectly recognized by machines based on different privacy protection requirements, we propose to incorporate some knowledge of facial recognition models into the denoising process for guidance. Inspired by training-free method \cite{yu2023freedom,kwon2022diffusion,avrahami2022blended,fei2023generative}, we construct an energy function to perform gradient correction on the diffusion model. According to \cite{yu2023freedom}, the formula can be written as follows:

\begin{equation}
    \nabla_{x_t}\log_{}{p(c|x_t)} \propto -\nabla_{x_t}\varepsilon(c,x_t),
    \label{eq:correction gradient}
\end{equation}

\begin{equation}
    x_{t-1}= x_t -\lambda_t \nabla_{x_t}\varepsilon(c,x_t),
    \label{eq:energy guide}
\end{equation}

where $x_t$ represents noisy image. $\lambda_t$ is a scale factor, which can be seen as the learning rate of the correction term.

Next, we need to construct the energy function to guide the denoising process. Classifier-based methods \cite{dhariwal2021diffusion,liu2023more,nichol2021glide,zhao2022egsde}  choose time-dependent distance measuring functions $D(c, x_t, t)$ to approximate the energy functions as follows: 
\begin{equation}
    \varepsilon(c,x_t) \approx D_\phi(c,x_t,t),
    \label{eq:classifier_based}
\end{equation}
where $\phi$ defines the pre-trained parameters. $D_\phi(c, x_t, t)$ computes the distance between the given condition $c$ and noisy image $x_t$. However, it is difficult to find an existing pre-trained network for noisy images. 

In order to leverage certain pre-trained models, the acquisition of clean images $\hat{x}_0$ from the noisy images $x_t$ of the diffusion process is necessary. Thanks to SDM is a latent diffusion model, we need to extract clean latent code $\hat{z}_0$ from the noisy latent code $z_t$. Referring to Eq.~\ref{eq:one_step nosing}, estimating the clean latent code $\hat{z}_0 $ from the noisy latent code $z_t$ becomes feasible. Subsequently, by employing the pre-trained decoder $D_{ec}$ of SDM, we can derive a clean image $\hat{x}_0$ as follows:

\begin{equation}
    \hat{x}_0 = D_{ec}(\hat{z}_0),
    \label{eq:predict x0}
\end{equation}

\begin{equation}
    \hat{z}_0 = \frac{z_t}{\sqrt{\alpha_t}}- \frac{\sqrt{1-\alpha_t}\epsilon_\theta(z_t,t,C_t)}{\sqrt{\alpha_t}}.
    \label{eq:predict z0}
\end{equation}

Accordingly, the energy function can be written as: $\varepsilon(c,x_t) \approx D_\theta(c,\hat{x}_0)$. Here, to mitigate symbol confusion, we will substitute the actual condition (original image $x$) instead of the condition $c$.

\begin{equation}
    \varepsilon(c,x_t) \approx D_\theta(x,\hat{x}_0).
    \label{eq:energy}
\end{equation} 

According to Eq.~\ref{eq:denosing}, Eq.~\ref{eq:correction gradient}, Eq.~\ref{eq:energy guide}, Eq.~\ref{eq:predict x0}, Eq.~\ref{eq:predict z0} and Eq.\ref{eq:energy}, the approximated sampling process can be writen as:

\begin{equation}
    z_{t-1}' = z_{t-1} - \lambda_t \nabla_{\hat{z}_0}D_\theta(x,\hat{x}_0),
    \label{eq:denosing_guidance}
\end{equation}

\begin{equation}
    z_{t-1}= \sqrt{\frac{\alpha_{t-1}}{\alpha_t}}z_t + \left(\sqrt{\frac{1}{\alpha_{t-1}}-1}-\sqrt{\frac{1}{\alpha_t}-1}\right)\cdot \epsilon_\theta\left(z_t,t,C_t\right).
    \label{eq:denosing1}
\end{equation}
This sampling process is reasonable because the clean latent code $\hat{z_0}$ has the same trend of change as the noisy latent code $z_t$. As the distance between the clean image $\hat{x}_0$ (decoded from the clean latent code $\hat{z}_0$) and the conditional image decreases, a corresponding reduction occurs in the distance between the noisy image $x_t$ (decoded from the noisy latent code $z_t$) and the conditional image.

\subsubsection{Anonymization}
\label{subsec:anonymization}
Anonymization aims to eliminate the identifiable features of individuals in pictures, ensuring that both human observers and machines cannot accurately recognize them. Simultaneously, the image quality and utility of anonymized faces are also important. The former enhances photo-realism and strengthens security, while the latter makes it possible to perform identity-agnostic tasks on an anonymized face image in a privacy-preserving manner. Based on the above objectives, we introduce the process of our method to achieve anonymization (as shown in Figure~\ref{fig:method_inference}).

\begin{wraptable}{r}{8cm}
\centering
\caption{\textbf{Quantitative evaluation for anonymization methods.} We calculate the successful protection rate (SR) for de-identification results and the identification rate for recovered results. A higher rate implies better performance.}
\begin{tabular}{cccc} 
\toprule
Type           & Method  & Facenet  & ArcFace  \\
\hline 
\multirow{5}{*}{De-identity$\uparrow$}         & Ours    & \textbf{0.988}    & \textbf{1}      \\
& RIDDLE    & 0.985    & \textbf{1} \\
& CIAGAN    & 0.849   & 0.9 \\
& RIDDLE    & 0.985    & 0.998 \\
& Deep Privacy    & 0.923    & 0.933 \\
\hline 
\multirow{3}{*}{Recovery$\uparrow$}             & Ours    & \textbf{1}    & \textbf{1}       \\
& RIDDLE    & \textbf{1}    & 0.8    \\
& FIT   &\textbf{1} & \textbf{1}   \\
\bottomrule
\end{tabular}
\label{tab:de-id}
\end{wraptable}

We first randomly add noise to the initial latent code $z_0$ according to Eq.~\ref{eq:one_step nosing}, where $t=S_{ns} \ast T$. $T$ is the total time step in the denoising process, $S_{ns}$ is a scaling factor that mainly controls the noise strength, where we take $S_{ns}=0.6$. Adding noise at this time step can maintain the overall layout of the image and disrupt the facial appearance of the image. Next, we will denoise the latent code $z_t$. Initially, we obtain the latent code $z_{t-1}$ of step $t-1$ using Eq.~\ref{eq:denosing1}. Subsequently, we calculate the latent code $z_{t-1}'$ of step $t-1$ with guidance using Eq.~\ref{eq:denosing_guidance}. Here, we use the identity dissimilarity loss ($L_{Idis}$) function and diversity loss ($L_{div}$) function to construct the energy function in Eq.~\ref{eq:denosing_guidance}. And we set $\lambda_t =1$ so that the loss function can converge to 0 during the denoising process. The identity dissimilarity loss function ensures that the generated image has a different identity from the original image.
\begin{equation}
    L_{Idis}= \sum_{i=1}^{4} Max(\frac{F_{\theta}(x) \cdot F_{\theta}(\hat{x}_0^i)}{||F(x)||\cdot ||F_{\theta}(\hat{x}_0^i)||},0),
    \label{eq:Idis loss}
\end{equation}
 where $\hat{x}_0^i$ is obtained by Eq.~\ref{eq:predict x0}, $F_{\theta}$ represents a pre-trained facial recognition model. Note that in order to enhance the diversity of generated results, we add different noises to the initial latent code $z_0$ and obtained different clean images $\hat{x}_0^i$ based on the Eq.~\ref{eq:one_step nosing}, Eq.~\ref{eq:denosing1} and Eq.~\ref{eq:predict x0}. The detailed process will be introduced below.
 
 Diversity loss is designed to enhance the diversity of anonymized faces. The intuition is to add different noises to the initial latent code (Eq.~\ref{eq:one_step nosing}), and the identity of the resulting anonymized face should also be inconsistent. In our experiment, four groups of different noises are added to each initial latent code $z_0$ (as shown in Figure.~\ref{fig:method_inference}). The diversity loss can be formulated as follows:
 \begin{equation}
     L_{div} =  \sum_{i=1}^{4} \sum_{j=2, j \neq i}^{4} Max(\frac{F_{\theta}(\hat{x}_0^i) \cdot F_{\theta}(\hat{x}_0^j)}{||F_{\theta}(\hat{x}_0^i)||\cdot ||F_{\theta}(\hat{x}_0^j)||},0).
     \label{eq:div loss}
 \end{equation}
 In addition to utilizing the identity guidance module, we develop an embedding scheduling strategy to implement anonymization more effectively. Specifically, we consider a two-stage embedding scheduling procedure, divided by $\tau \in [0,1]$. For a denoising process with $T$ steps, one can employ unconditional embedding $UC$ at the first $\tau \ast T$ steps, and use the learned embedding $C^{t/100}$ corresponding to each step for the remaining $(1-\tau) \ast T$ steps. Here, $\tau$ is set to $0.4$.

\begin{equation}
     \begin{split}
         C_t = \left\{
         \begin{array}{ll}
            C^{t/100}  & t> \tau \ast T \\
            UC  & t\leq \tau \ast T
         \end{array}
         \right.
     \end{split}
     \label{eq:emb_strategy1}
 \end{equation}

\begin{table}[t]
\centering
\caption{Utility evaluation of de-identification results.} 
\begin{tabular}{c|c|ccccc} 
\toprule
\multicolumn{2}{c|}{Method}           & Ours  & RIDDLE  & CIAGAN & FIT & DeepPrivacy  \\
\hline 
\multirow{2}{*}{Face detection $\uparrow$}  & MtCNN   & \textbf{1}    & \textbf{1}    & \textbf{1}  & \textbf{1} & \textbf{1}      \\
 & Dlib  & \textbf{1}    & \textbf{1}    & 0.957  & \textbf{1} & 0.995 \\
 \hline
 \multirow{2}{*}{Bounding box distance $\downarrow$}  & MtCNN   & 5.834    & 6.45    & 19.975  & \textbf{5.3} & 6.467   \\ 
  & Dlib  & \textbf{3.695}    & 6.245    & 18.441  & 3.866 & 5.296 \\
  \hline
\multirow{2}{*}{Landmark distance $\downarrow$}  & MtCNN   & 2.848    & 3.478    & 7.326  & \textbf{2.398} & 4.298   \\ 
  & Dlib  & \textbf{2.548}    & 3.453    & 9.598  & 2.664 & 4.075 \\
\bottomrule
\end{tabular}
\label{tab:utility}
\end{table}

\begin{figure}[t]
\centering
\includegraphics[width=1.0\linewidth]{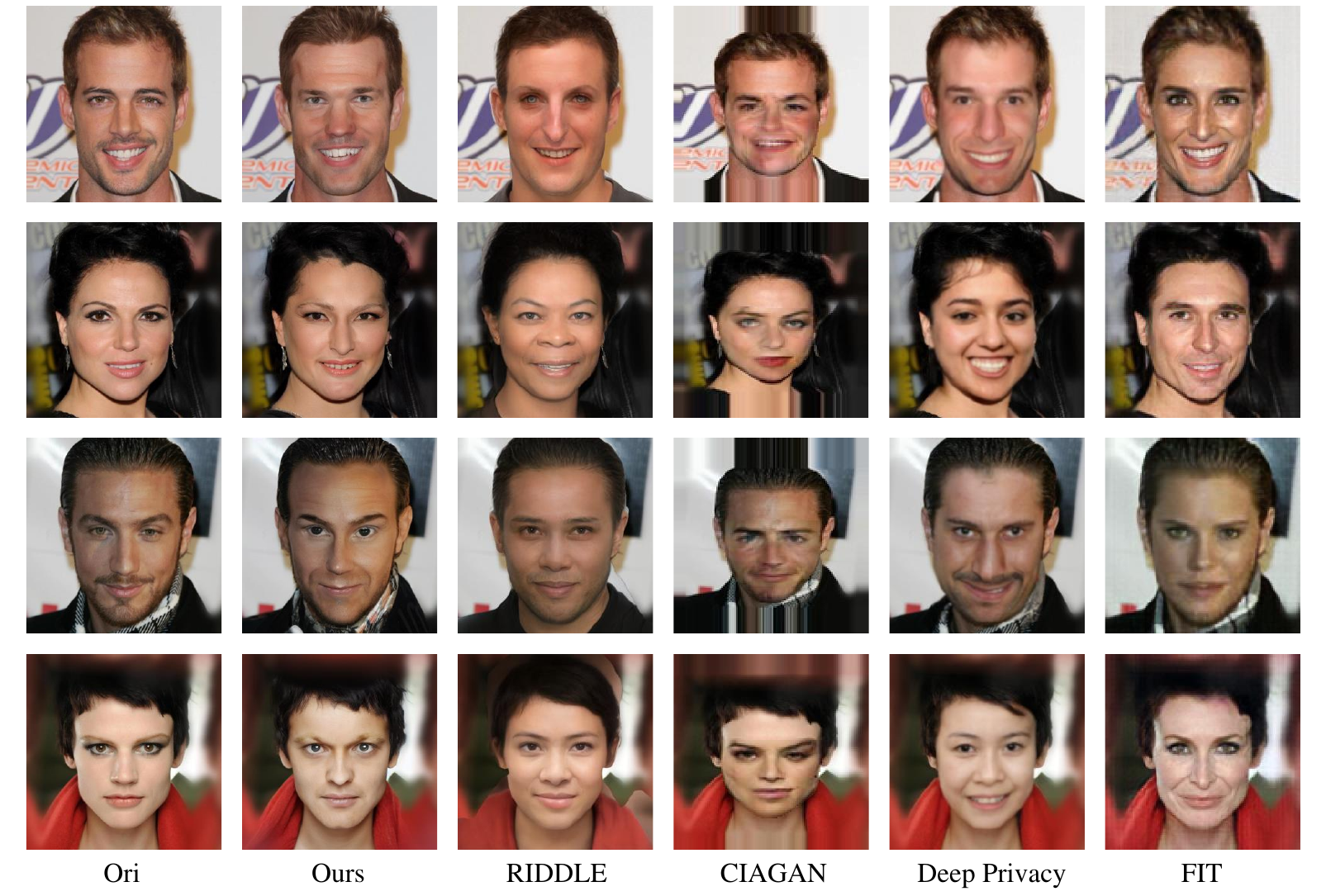}
\caption{\textbf{Qualitative comparison with literature anonymization methods.}}
\label{fig:anony_deid}
\end{figure}
 
This strategy enables us to achieve anonymization while preserving the overall layout of the original image and certain identity-independent attributes (e.g., posture) unchanged. Furthermore, in order to recover the image better, we also generated a noisy latent $z_T$ as key-I through DDIM Inversion (Eq.~\ref{eq:nosing}) in the process of anonymization, where the conditional embedding $C = C^{t/100}$.

\subsubsection{Visual identity information hiding}
\label{subsec:Visual information hiding}

Visual identity information hiding aims to generate images that cannot be recognized by humans but can be accurately recognized by machines. Compared with anonymization, visual identity information hiding does not require the generated image to maintain the pose and background unchanged. However, we still aim to generate a photo-realism facial image. Indeed, this setting significantly enhances the security of privacy protection. In the event of a data breach, hackers will be uncertain whether the generated faces are real or synthetic, making it much more challenging for them to compromise individual privacy. Based on the above objectives, we introduce the process of our method to achieve visual identity information hiding.

\begin{figure}[htbp]
\centering
\begin{subfigure}{0.49\linewidth}
    \centering
    \includegraphics[width=1.\linewidth]{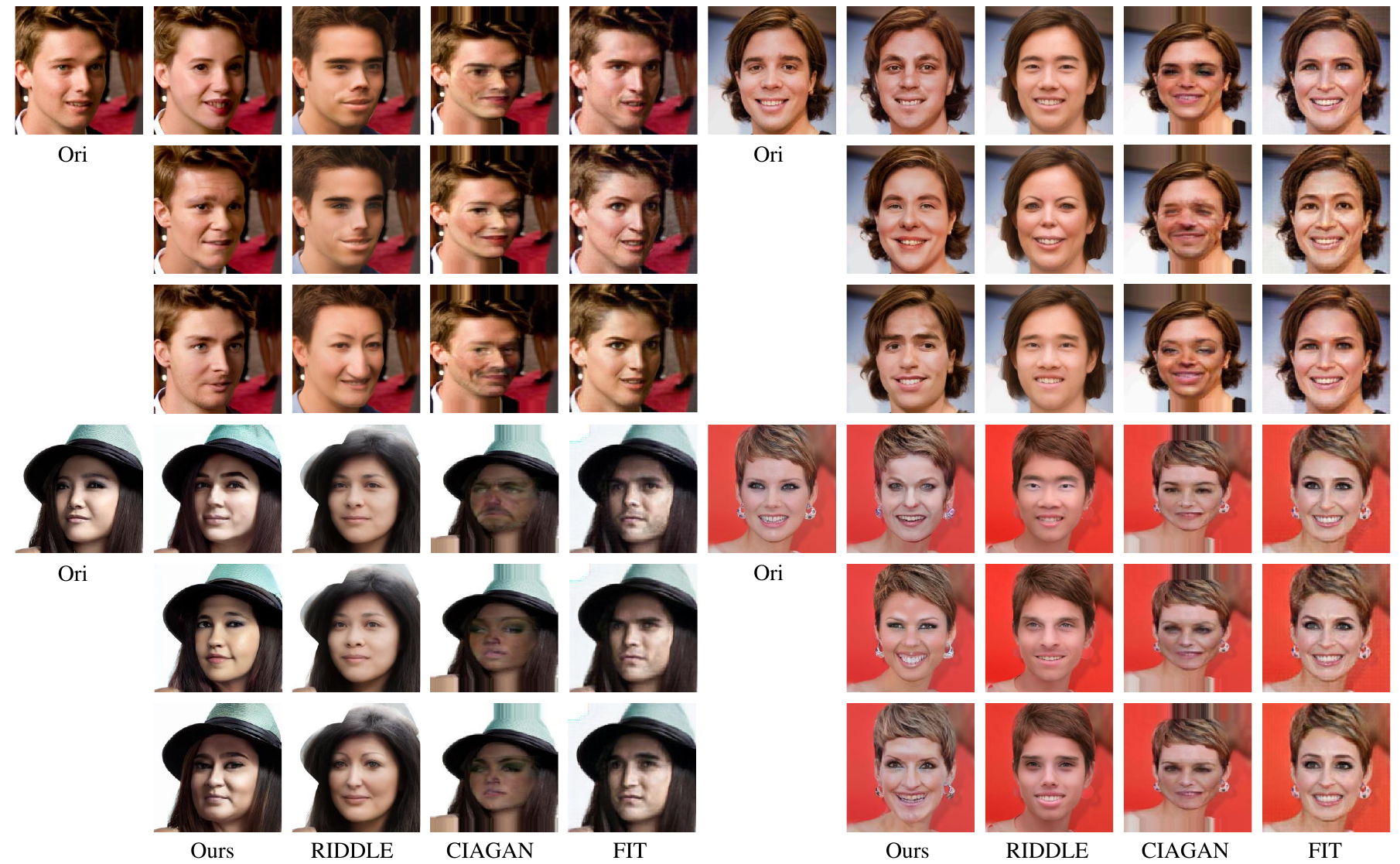}
    \caption{Qualitative comparison}
    \label{diversity1}
\end{subfigure}
\centering
\begin{subfigure}{0.49\linewidth}
    \centering
    \includegraphics[width=1.\linewidth]{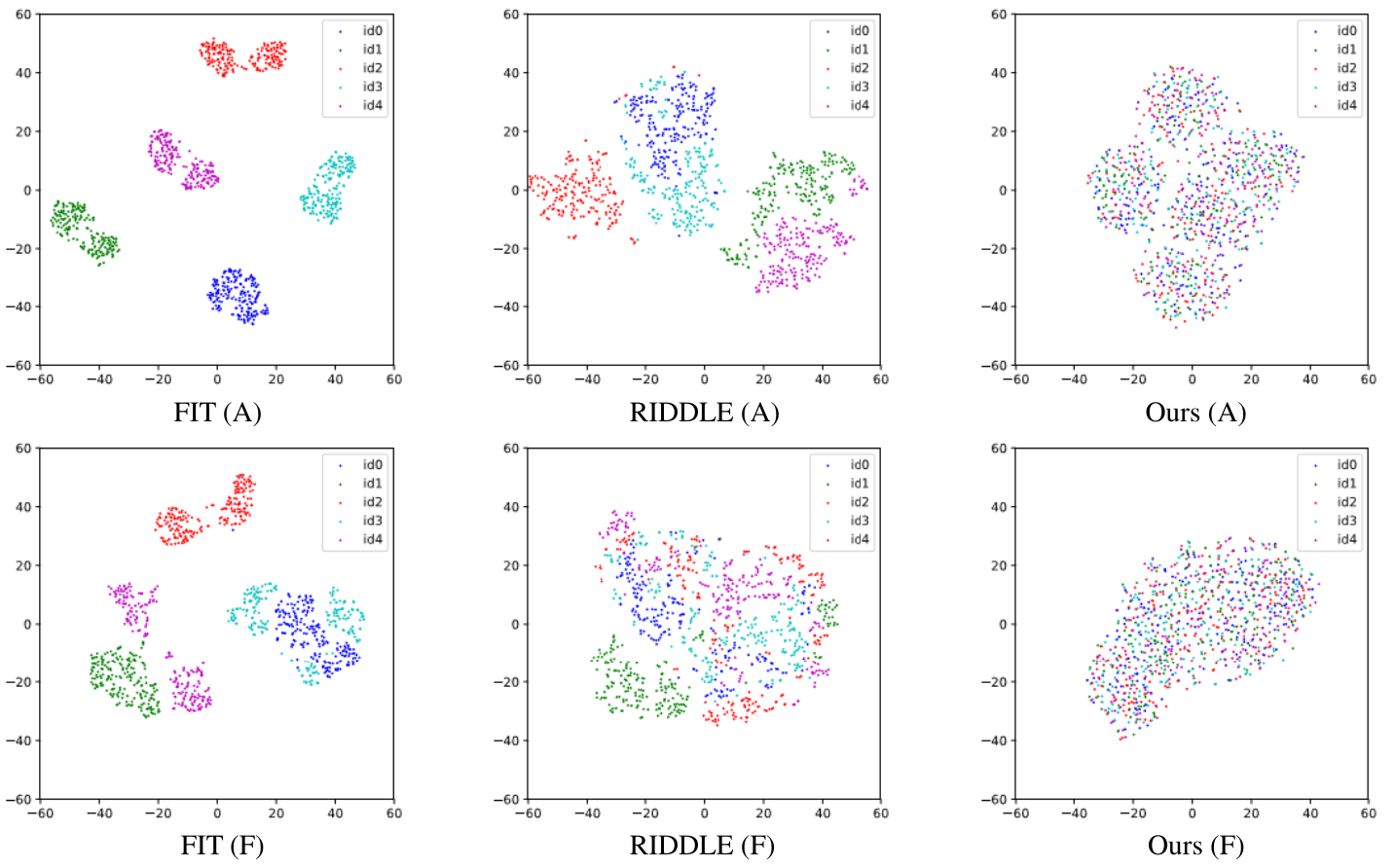}
    \caption{Quantitative comparison}
    \label{diversity2}    
\end{subfigure}
\caption{ Qualitative and quantitative comparison with literature anonymization methods on diversity. }
\label{fig:diversity}
\end{figure}

The overall process of visual identity information hiding is similar to anonymization. Initially, we employ Eq.~\ref{eq:one_step nosing} to introduce noise to the initial latent code $z_0$ and obtain the noisy latent code $z_t$. In this context, $S_{ns}$ is set to $0.8$ to ensure a considerable change in the perspective perceived by human observers, including layout and background. Next, we also obtain the latent code $z_{t-1}$ of step $t-1$ Eq.~\ref{eq:denosing1}, and the latent code $z_{t-1}$ of step $t-1$ with guidance through Eq.~\ref{eq:denosing_guidance}. The energy function in Eq.~\ref{eq:denosing_guidance} is the identity similarity loss $L_{is}$. We set different $\lambda_t$ for different recognition networks. In principle, we select the value that can make the cosine similarity between the encrypted and original images in the recognition embedding space around 0.95. Identity similarity loss promotes the generated image that can be correctly recognized by machines, which can be formulated as:
 \begin{equation}
     L_{is} = 1- \frac{F_{\theta}(x) \cdot F_{\theta}(\hat{x_0})}{||F_{\theta}(x)||\cdot ||F_{\theta}(\hat{x_0})||}.
     \label{eq: is loss}
 \end{equation}

Regarding the scheduling strategy for embedding, as we only need to constrain the image to a facial manifold without maintaining some attributes such as posture, we set $\tau$ in Eq.~\ref{eq:emb_strategy1} to $0.6$. In addition, in order to recover the original image, we also use DDIM Inversion (Eq.~\ref{eq:nosing}) as described in Section~\ref{subsec:anonymization} to generate a noisy latent code $z_T$ as the key-I.

\subsection{Identity recovery}
\label{sec:recover}
 In the previous section, we have implemented facial privacy protection and obtained key-E and key-I. Now, we use key-I as the noisy latent code $z_T$ and key-E as conditional embedding for denoising to recover the original image. Specifically, through Eq.~\ref{eq:denosing1}, we can gradually denoise the noisy latent code $z_T$ to obtain the initial latent code ${z_0}^{'}$, and decode it to obtain the recovered image $x^{'}$, where $C_t = C^{t/100}$.

\section{Experiments and evaluation}

\subsection{Implementation details}
\begin{table}[t]
\centering
\caption{Quantitative comparison on recovered results. AVIH (F) and AVIH (H) represent AVIH using FaceNet and ArcFace as target models, respectively.} 
\begin{tabular}{ccccccc} 
\toprule
\multirow{2}{*}{Method}    & \multicolumn{3}{c}{Anonymization} &\multicolumn{3}{c}{Visual identity information hiding}  \\
\cmidrule(r){2-4}  \cmidrule(r){5-7}
& FIT & RIDDLE & Ours  & AVIH (F) & AVIH (A) & Ours \\
\hline 
 MSE $\downarrow$     & 0.006    & 0.045    & \textbf{0.003}   & \textbf{0.003} & 0.004 & 0.004      \\
 LPIPS$\downarrow$    & 0.045    & 0.192    & \textbf{0.494}   & 0.216 & 0.109 & \textbf{0.059}  \\
 SSIM$\uparrow$       & 0.762    & 0.494    & \textbf{0.854}   & 0.775 & 0.793 & \textbf{0.872}  \\ 
 PSNR$\uparrow$       & 28.693   & 19.489   & \textbf{28.900}  &\textbf{32.306} &31.369 &31.913 \\
\bottomrule
\end{tabular}
\label{tab:recover}
\vspace{-2mm}
\end{table}
We have retained the original hyperparameter selection of SDM, and only the parameters of our proposed MSI module can be trained. On an NVIDIA GeForce RTX3090, the training process for each image takes approximately $20$ minutes, with a batch size of $1$. The basic learning rate is set to $0.001$. Next, we will evaluate the effectiveness of our algorithm on the CelebA-HQ \cite{karras2017progressive} dataset and LFW \cite{huang2008labeled} dataset.

\subsection{Anonymization}

We compare our method with the anonymization methods: RiDDLE \cite{li2023riddle}, FIT \cite{gu2020password}, CIAGAN \cite{maximov2020ciagan} and DeepPrivacy \cite{hukkelaas2019deepprivacy}. Note that RiDDLE and FIT can achieve recoverable anonymization. Next, we will evaluate the performance of our method in both anonymization and recovery.

\begin{figure}[t]
\centering
\includegraphics[width=1.0\linewidth]{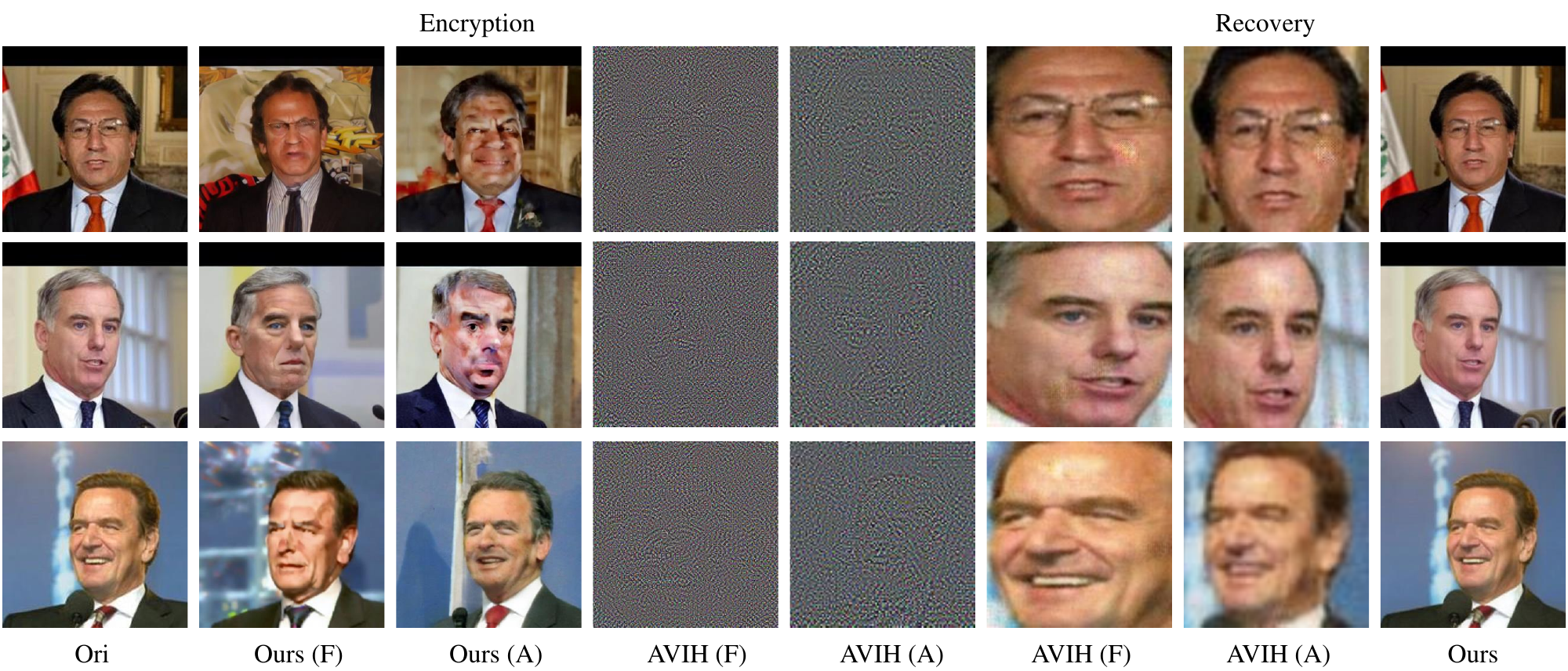}
\caption{\textbf{Qualitative comparison with literature visual identity information methods.}}
\label{fig:second_results}
\vspace{-2mm}
\end{figure}

\subsubsection{De-identification}
\label{subsubsec:exp_anoy_deid}
\begin{wrapfigure}{r}{7cm}
\centering
\includegraphics[width=1.0\linewidth]{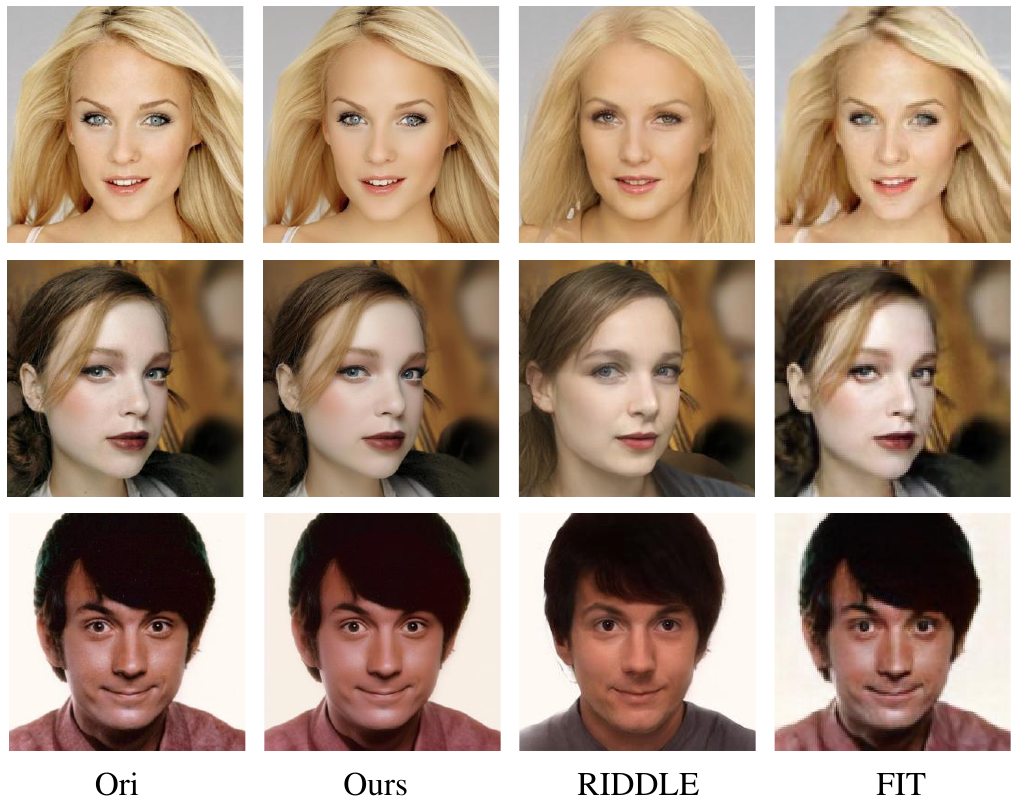}
\caption{\textbf{Visualization comparison of recovered image between different anonymization methods.} Zoom in for a better view.}
\label{fig:anony_recover}
\vspace{-4mm}
\end{wrapfigure}
The qualitative comparison is shown in Figure~\ref{fig:anony_deid}. CIAGAN \cite{maximov2020ciagan} experienced significant distortion when generating anonymized images. FIT \cite{gu2020password} can generate faces with significantly different identities from the original image. However, its visual quality is unsatisfactory, and the generated faces do not match other attributes in the image (as shown in the second line of Figure~\ref{fig:anony_deid}, which generates a middle-aged man's face in a face that is clearly female). Riddle can generate diverse faces, but some important parts of the generated faces are unnatural, such as the eyes in the first row of Figure~\ref{fig:anony_deid}. Deepprivacy \cite{hukkelaas2019deepprivacy} can somewhat maintain photo-realism but fails to retain the identity-irrelevant attributes such as expressions. Compared with the above methods, our method generates images with natural facial features and better photo-realism. Moreover, our method can maintain identity-irrelevant attributes such as expressions and poses.
\begin{wraptable}{r}{7cm}
\centering
\caption{Identification rate comparison. A higher identification rate implies better performance.} 
\begin{tabular}{cccc} 
\toprule
Method           & Original  & Ours  &AVIH  \\
\hline 
 FaceNet       & 84.15\%    & \textbf{81.10\%}    & 78.52\%       \\
 ArcFace    & 88.30\%    & 87.70\%    & \textbf{88.30\%}    \\ 
 Average    & 86.23\%    & \textbf{84.40\%}    & 83.41\%    \\ 
\bottomrule
\end{tabular}
\label{tab:encrypt_visual}
\end{wraptable}
Quantitatively, we calculate the successful protection rate (SR) of different methods on the CelebA dataset. Note that when the distance between the identity embedding of the de-identified image and the source image exceeds the threshold set by the corresponding facial recognition network, protection is considered successful. Here, we use FaceNet and ArcFace for evaluation and choose the threshold of ArcFace as 0.8 and the FaceNet as 1.1 according to \cite{schroff2015facenet}. Table~\ref{tab:de-id} shows that our method has a higher SR than other anonymization methods, which proves that our method has better security.

\textbf{Face Utility.} In Table~\ref{tab:utility}, we apply computer vision algorithms on the de-identified images and evaluate the utility of de-identified images on downstream vision tasks. We calculate the face-detection rate between our methods and other anonymization methods on two face-detection models: MtCNN \cite{zhang2016joint} and Dlib \cite{kazemi2014one}. The per-pixel distance of facial bounding boxes and 68 facial key points are also calculated. As shown in the table, our method achieved the best results in face detection rate and pixel distance calculated based on the Dlib model and the comparable results as FIT in pixel distance calculated based on the MtCNN model. It means that our method can guarantee the consistency of the face region and landmarks better and better utility for identity-agnostic computer vision tasks.

\textbf{Diversity of identities.} We compared our method with FIT, RIDDLE, and CIAGAN for the diversity of de-identified images. The results are shown in Figure.\ref{fig:diversity}. Although all these methods can generate diversity faces, the faces generated by FIT under different passwords exhibit shared characteristics in local regions. CIAGAN generates faces with low-quality and obvious splicing traces. RIDDLE can obtain diverse faces, but the facial features obtained by encrypting different faces with the same password are similar. In contrast, our method brings fruitful facial features with high-quality for different images. To further demonstrate the diversity of our method, we conduct an identity visualization experiment. For each person, we use 200 different passwords or labels to generate their de-identified faces. Then, we use a face recognition network to extract identity embeddings of de-identified faces and perform dimensionality reduction using t-SNE \cite{van2008visualizing}. From Figure.~\ref{diversity2}, it can be seen that the identity clusters of FIT are relatively tight, and different clusters are spaced far apart on the hyperplane. In contrast, our de-identified faces are more dispersed and occupy most of the area of the hyperplane. 

\subsubsection{Identity recovery}
In this section, we compare our method with recoverable anonymization methods FIT and RiDDLE regarding identity recovery performance. We first calculate the identification rate of the recovered image (as shown in the bottom part of Table~\ref{tab:de-id}), where a higher identification rate indicates better proof of identity recovery. Then, we calculate the similarity between the original image and the recovered image. Specifically, we use mean square error (MSE), peak signal-to-noise ratio (PSNR), structural similarity (SSIM), and learned perceptual image patch similarity (LPIPS) as our metrics. From Table~\ref{tab:recover}), it can be seen that our method outperforms the existing recoverable anonymization methods in terms of identity recovery. In addition, as shown in Figure~\ref{fig:anony_recover}, we can see that compared with FIT and RIDDLE, the recovered image generated by our method is smoother, clearer, and more similar to the original image.

\begin{table}
\centering
\caption{The effectiveness of a set of conditional embeddings in identity recovery.} 
\begin{tabular}{c|c|ccccc} 
\toprule
\multicolumn{2}{c|}{Method} & MSE$\downarrow$  & LPIPS$\downarrow$  & SSIM$\uparrow$ & PSNR$\uparrow$  \\
\hline 
\multirow{2}{*}{Anonymization}  & Ours-OE   &  0.007   & 0.051    & 0.821  & 27.359 &     \\
 & Ours  & \textbf{0.003}    & \textbf{0.037}    & \textbf{0.854}  & \textbf{28.900} \\
 \hline
 \multirow{2}{*}{Visual identity information hiding}  & Ours-OE  & 0.005    & 0.080    & 0.831  & 30.237   \\ 
  & Ours  & \textbf{0.004}    & \textbf{0.059}    & \textbf{0.872}  & \textbf{31.913} \\
\bottomrule
\end{tabular}
\label{tab:ablation_OE}
\end{table}

\subsection{Visual identity information hiding}

In this section, we will compare our method with the recoverable visual identity information hiding method AVIH \cite{Su2022VisualIH}. AVIH exploits a generative model pre-trained in a private training setting as the key model and optimizes the protected image online based on the key model and pre-trained face recognition model. Next, we will evaluate the methods from two aspects: information encryption and identity recovery. 

\subsubsection{Information encryption}
We evaluate our approach according to the testing process proposed by AVIH. We randomly select 12 individuals from the LFW dataset as the probe set $Set_p$ and randomly select 10 images for each individual from the probe set as the same-identity verification set $Set_s$. A total of 12878 images from other individuals are used as the different-identity verification set $Set_d$. Then, we encrypt the images of same-identity verification set $Set_s$. In the evaluation stage, we sequentially take a face image from the probe set as a query and calculate the feature similarity between the query and images of the same identity in $Set_s$, as well as the feature similarity between the query and the images in $Set_d$. If there are images in $Set_d$ with higher similarity than the images in $Set_s$, it is considered that the recognition is incorrect. 
\begin{figure}
\centering
\includegraphics[width=1.0\linewidth]{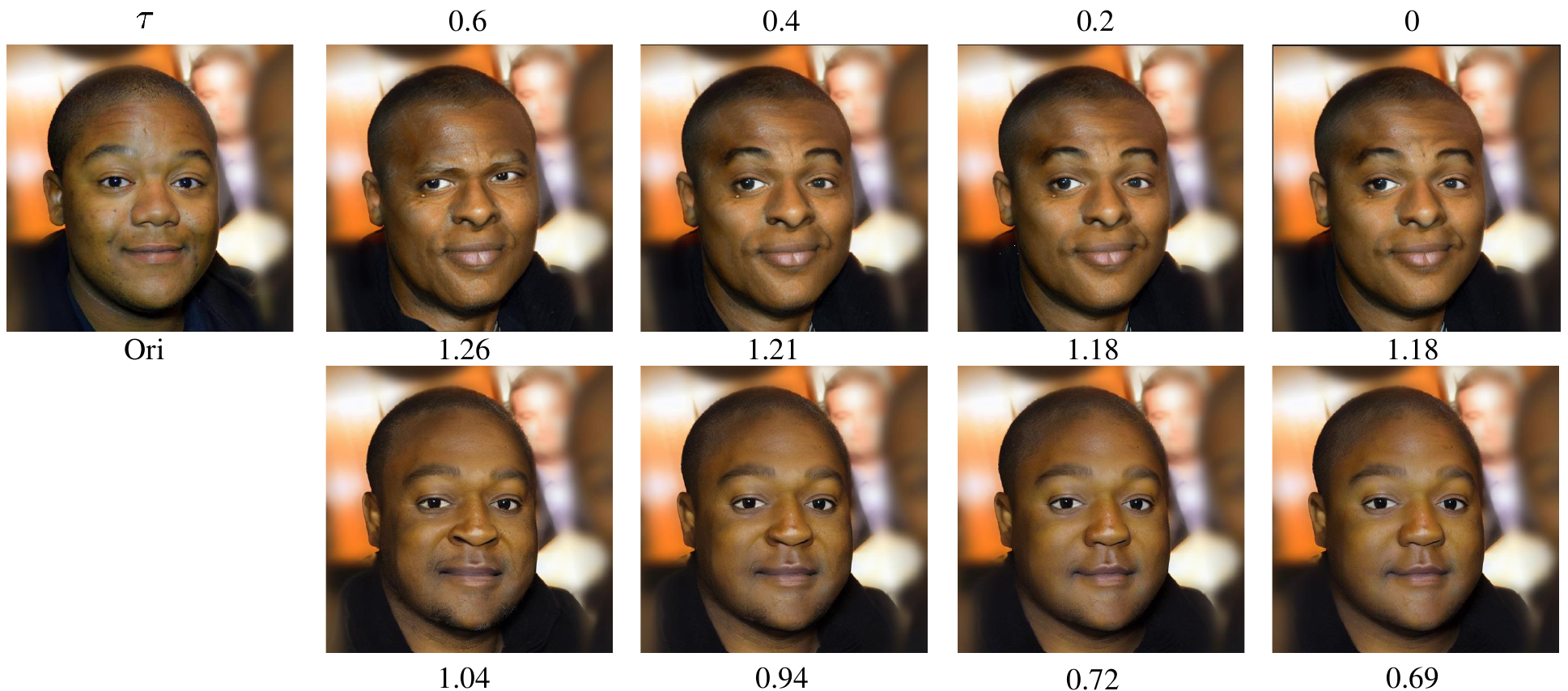}
\caption{\textbf{Effect of different conditional embedding in our method.} The first line of the image is generated using a set of conditional embeddings, while the second line is generated using one conditional embedding. The images in each row, arranged from left to right, represent the outcomes produced by the different $\tau$ of the embedding scheduling strategy. The number below the image represents the distance between the identity embedding of the de-identified face and the identity embedding of the original face.}
\label{fig:ablation_OE}
\end{figure}

The results are shown in Table~\ref{tab:encrypt_visual}. When using FaceNet as the face recognition model, our method has a recognition accuracy of three percentage points higher than AVIH and only three percentage points lower than the recognition rate of real images. Meanwhile, when utilizing the ArcFace face recognition model, our method exhibited comparable performance to AVIH, approaching the recognition accuracy achieved with real images. The results can demonstrate the utility of our method on the face recognition models in practical applications. In addition, the qualitative results are shown in Figure~\ref{fig:second_results}. It can be seen that our method encrypts the original image while preserving the facial structure in a photo-realistic manner. A notable advantage over AVIH is that even if a hacker invades the server, distinguishing whether these images have been encrypted becomes challenging, thus enhancing privacy protection security.

\subsubsection{Identity recovery}
In this section, we will evaluate the quality of the decrypted image and its similarity to the source image. The results are shown in Figure~\ref{fig:second_results} and Table~\ref{tab:recover}. Although AVIH achieved comparable performance to our method in the metrics of MSE and PSNR by pixel-level optimization of the image, our method outperformed AVIH in the perceptual-level metrics of LPIPS and SSIM. Moreover, Figure~\ref{fig:second_results} illustrates that AVIH generates recovered images with artifacts and can only recover the image content of the area where the face is located. In contrast, our method can achieve complete recovery of the original image in a high-quality manner.

\begin{figure}
\centering
\includegraphics[width=1.0\linewidth]{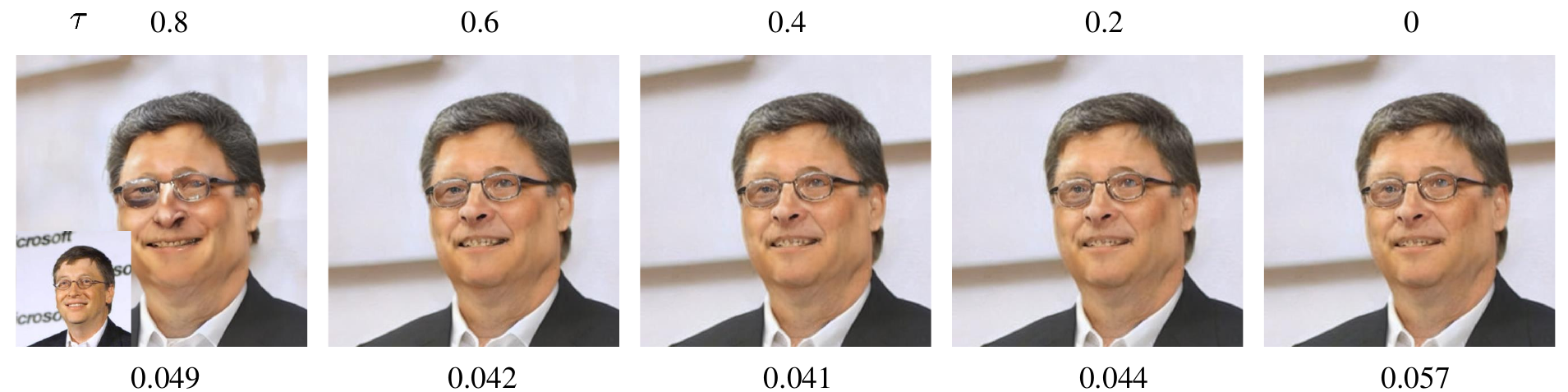}
\caption{\textbf{Visualization of generated results of visual identity information hiding tasks under different embedding scheduling strategies.} The number below the image represents the identity embedding distance between the encrypted face and the original face.}
\label{fig:ablation_strategy}
\end{figure}

\begin{figure}
\centering
\includegraphics[width=1.0\linewidth]{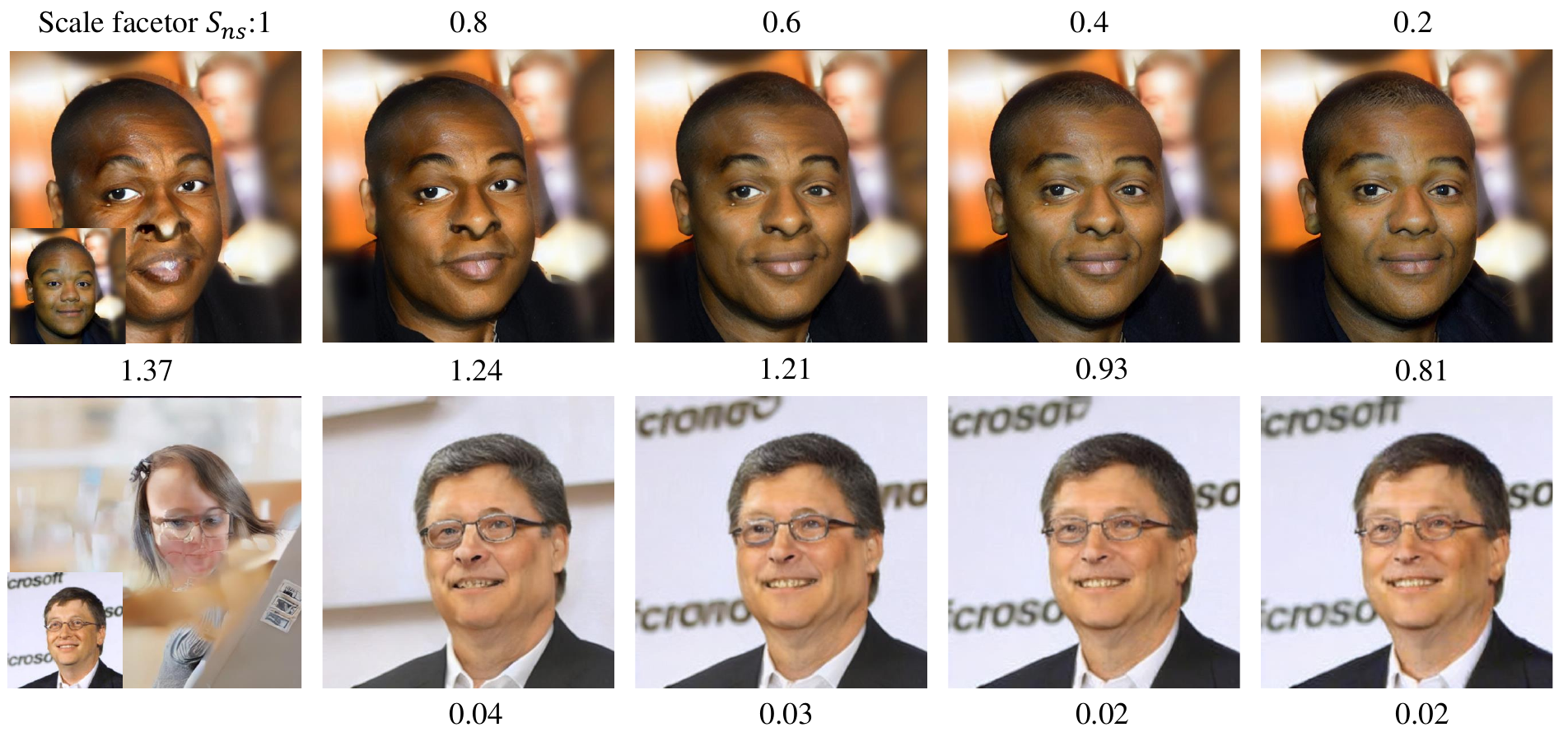}
\caption{\textbf{Effect of different noise strength in our method.} The number below the image represents the distance between the identity embedding of the encrypted face and the identity embedding of the original face. Because the image in the second row and first column can not detect a face, the distance between it and the original face in the recognition embedding space is not given.}
\label{fig:ablation_strength}
\end{figure}

\subsection{Ablation study}
In this section, we will evaluate the contribution of each component in our model.
\begin{wrapfigure}{r}{7cm}
\centering
\includegraphics[width=1.0\linewidth]{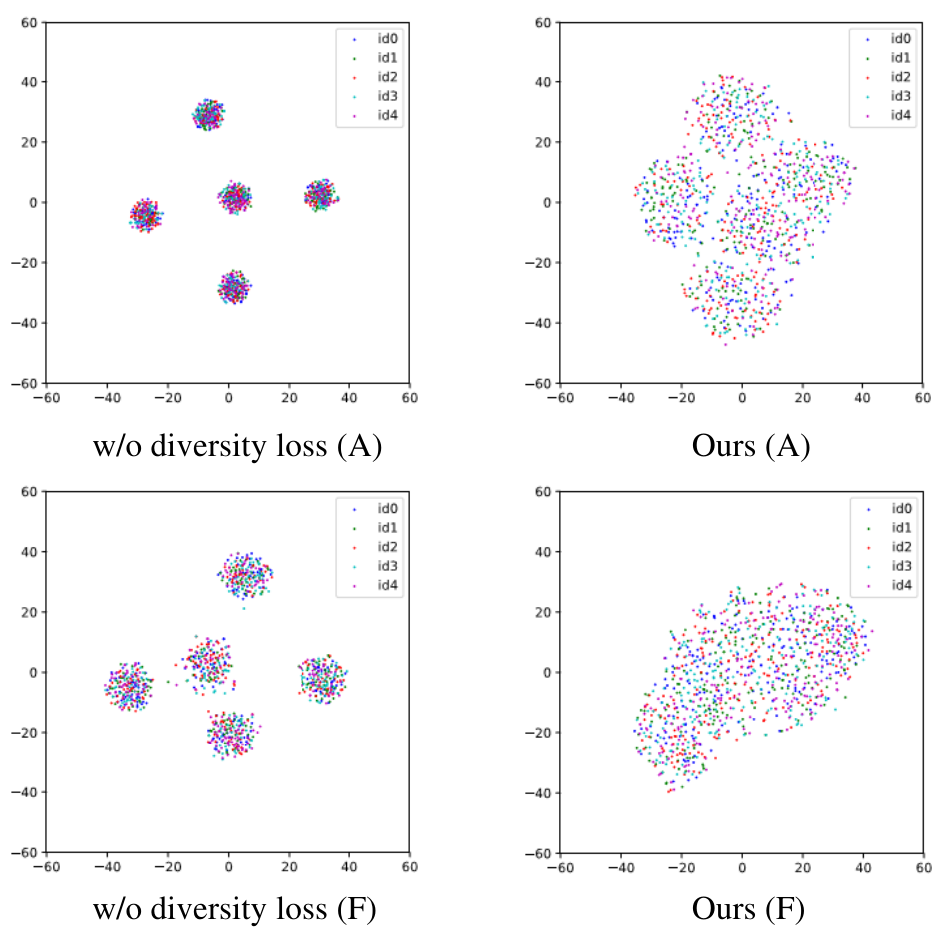}
\caption{\textbf{The effectiveness of diversity loss function.} We remove the diversity loss and conducted experiments to verify its effectiveness.}
\label{fig:ablation_diversity}
\end{wrapfigure}
\textbf{A set of conditional embeddings.}
In our method, we design an MSI module to obtain a set of conditional embeddings of the original image. Now, we conduct experiments to verify the advantages of a set of conditional embeddings over one conditional embedding. Specifically, we remove the multi-layer features and time modulation modules from the MSI module and only use the last layer of features encoded by clip to obtain one conditional embedding, represented as \textbf{Ours-OE}. Figure.~\ref{fig:ablation_OE} shows the de-identified results of using one embedding and a set of conditional embeddings in anonymization tasks. It can be seen that using a set of conditional embeddings has better decoupling and editability, and different levels of privacy protection can be achieved by using different embedding scheduling strategies. On the contrary, using one conditional embedding has poor editability and generates unsatisfactory de-identified faces. In addition, we also conducted experiments on the impact of one embedding and a set of embeddings on identity recovery. The results are shown in Table.~\ref{tab:ablation_OE}. From the table, it can be seen that using a set of embeddings for identity recovery outperforms using a single embedding in both pixel-level and perceptual-level metrics. The above experiment demonstrates the effectiveness of using a set of conditional embeddings for encryption and recovery.

\textbf{Embedding scheduling strategy.}
We have designed corresponding embedding scheduling strategies for privacy protection tasks with different requirements. Next, we conduct experiments to verify the effectiveness of these strategies and provide the impact of different strategies on the generated results. Specifically, we conduct experimental verification by changing $\tau$ in the embedding scheduling strategy. The results are shown in Figure.~\ref{fig:ablation_OE} and Figure.\ref{fig:ablation_strategy}. From the first line of Figure.~\ref{fig:ablation_OE}, it can be seen that as $\tau$ decreases, the similarity between the de-identified face and the original face gradually increases, and the identity embedding distance between the de-identified face and the original face decreases. When $\tau=0.4$, it ensures the effectiveness of anonymization while maintaining irrelevant attributes such as posture and facial expressions unchanged. In addition, it can be seen from Figure.\ref{fig:ablation_strategy} that in the visual identity information hiding task, when $\tau=0.4-0.6$, the distance between the encrypted face and the original face in the face recognition embedding space is the smallest. In our method, in order to ensure significant changes in the identity of encrypted images from the perspective of human observers, we chose the time embedding scheduling strategy with $\tau=0.6$.

\textbf{Noise strength.}
We change the scaling factor $S_{ns}$ when adding noise to explore the impact of different noise strengths on the generated results. As shown in Figure.~\ref{fig:ablation_strength}, when the noise strength weakens (i.e., the scaling factor $S_{ns}$ decreases), the generated results gradually resemble the original image. In the first row of the figure (anonymization task), it can be seen that when $S_{ns}=0.6$, privacy protection is effectively achieved while ensuring that identity-independent attributes such as posture and facial expressions remain unchanged. In the visual identity information hiding task (the second line in the figure), when $S_{ns}=0.8$, it maximizes the difference observed by the naked eye while ensuring the correct recognition of facial identity by the machine.

\textbf{Diversity loss function.}
In the anonymization task, in order to enhance the diversity of de-identified images, we design a diversity loss function. Now, we remove the diversity loss function and conduct experiments to verify its effectiveness. This variant is called \textbf{w/o diversity loss}. We conduct identity visualization experiments as described in Section~\ref{subsubsec:exp_anoy_deid}. The results are shown in Figure.~\ref{fig:ablation_diversity}. With diversity loss, the de-identified faces are more scattered and occupy most of the area of the hyperplane. 
\section{Conclusion}

In this paper, we unify anonymization and visual identity information hiding tasks and propose a novel face privacy protection method based on diffusion models, dubbed Diff-Privacy. It mainly achieves recoverable facial privacy protection through three stages. Stage I. Learning a set of conditional embedding as the key-E through our proposed MSI module. Stage II. Accomplishing privacy protection through our energy function-based identity guidance module and embedding scheduling strategy during the denoising process and then getting a noised map as the key-I according to DDIM inversion. Stage III. Performing identity recovery using DDIM sampling based on the acquired key. Extensive experiments demonstrate our method achieves state-of-the-art results both quantitatively and qualitatively.


%
{\small
\bibliographystyle{ieee_fullname}
\bibliography{main}
}


\end{document}